\documentclass{midl} 

\usepackage{mwe} 

\usepackage{array,multirow,graphicx}
\usepackage{lipsum}
\usepackage{pifont}
\usepackage{mathtools}
\usepackage{xcolor}

\usepackage[font=small,skip=0pt]{caption}
\usepackage{titlesec}
\titlespacing\section{0pt}{8pt}{4pt}

\definecolor{rcs}{RGB}{40,0,215}
\definecolor{rcn}{RGB}{83,0,172}
\definecolor{rrs}{RGB}{127,0,128}
\definecolor{rrn}{RGB}{78,0,177}
\definecolor{rcrs}{RGB}{83,0,172}
\definecolor{rcrn}{RGB}{81,0,174}

\newcommand{\coord}{\textit{coord}}
\newcommand{\Coord}{\textit{Coord}}
\newcommand{\conv}{\textit{conv}}
\newcommand{\Conv}{\textit{Conv}}

\newcommand{\bb}[1]{\textbf{#1}}

\let\clearpage\relax

\jmlrproceedings{MIDL}{Medical Imaging with Deep Learning}
\jmlrpages{}
\jmlryear{2022}

\jmlrvolume{-}

\title[Scale-Agnostic Super-Resolution in MRI using Coordinate Networks]{Scale-Agnostic Super-Resolution in MRI using Feature-Based Coordinate Networks}

\midlauthor{\Name{Dave {Van Veen}}, \Name{Rogier {van der Sluijs}}, \Name{Batu Ozturkler}, \Name{Arjun Desai}, \Name{Christian Bluethgen}, \Name{Robert D.~Boutin}, \Name{Marc H. Willis}, \Name{Gordon Wetzstein}, \Name{David Lindell}, \Name{Shreyas Vasanawala}, \Name{John Pauly}, \Name{Akshay S.~Chaudhari} \\
\Email{\{vanveen, sluijs, ozt, arjundd, bluch, boutin, mwillis7, \\ gordonwz, lindell, vasanawala, pauly, akshaysc\}@stanford.edu}
\\
\addr Stanford University
}

\begin{document}

\maketitle

\begin{abstract}
We propose using a coordinate network decoder for the task of super-resolution in MRI. The continuous signal representation of coordinate networks enables this approach to be scale-agnostic, i.e.~one can train over a continuous range of scales and subsequently query at arbitrary resolutions. 
Due to the difficulty of performing super-resolution on inherently noisy data, we analyze network behavior under multiple denoising strategies. Lastly we compare this method to a standard convolutional decoder using both quantitative metrics and a radiologist study implemented in Voxel\footnote{\url{https://voxel.im}}, our newly developed tool for web-based evaluation of medical images.
\end{abstract}

\begin{keywords}
Coordinate networks, super-resolution, MRI.
\end{keywords}

\vspace{7mm}
\section{Introduction}
High-resolution scans in magnetic resonance imaging (MRI) to depict high-frequency details such as local textures and edges are crucial for many diagnostic imaging tasks.
However, tradeoffs with scan time and signal-to-noise ratios motivate improved MRI resolution for higher downstream clinical utility.
While deep learning can leverage data-driven priors for encoding high-frequency information in super-resolution tasks, state-of-the-art methods are limited to upsampling at fixed, discrete scales due to their convolutional structure~\cite{lim2017enhanced, chen2021pre}.
Discrete scales are undesirable for clinical interpretation~\cite{chaudhari2021prospective}. Further, training such fixed networks places strict limits on acquiring homogeneous training data. In clinical settings, resolution for MRI acquisitions often varies based on factors such as patient size~\cite{wargo2013comparison}. Hence to construct a training dataset with uniform resolution, individual images must be resized, which can lead to poorer model performance~\cite{karras2019style, chai2022any}.

We propose a \textit{scale-agnostic} framework for MRI super-resolution using a coordinate network as a decoder. Given a coordinate, this decoder queries the neighboring latent features to predict the pixel value at that location. The continuous nature of this decoder allows querying at arbitrary resolutions; it also decouples the training and querying scales, e.g.~one can train on a continuous range of 1-2$\times$ and query at 3$\times$. Additionally, since super-resolution tasks are prone to augment noise~\cite{singh2014super}, we analyze the behavior of coordinate networks with various denoising strategies. This demonstrates the importance of early stopping, a technique commonly used in convolutional networks to avoid overfitting~\cite{heckel2019denoising}. Finally we compare the proposed framework's coordinate decoder against a standard convolutional decoder, using image quality metrics and a clinical reader study.

\subsection*{Contributions}

\begin{itemize}
    \itemsep0em
    \item We propose a framework for MRI super-resolution that is \textit{scale-agnostic}. This enables both decoupling between training and querying scales and also querying at arbitrary resolutions.
    \item We demonstrate various regularization strategies for coordinate networks in the presence of noisy data, including early stopping and the use of denoisers during network training.
    \item In addition to quantitative metrics, we evaluate results with a radiologist reader study built on Voxel, our newly developed tool for web-based evaluation of medical images.
\end{itemize}
\section{Related Work}
Coordinate networks, sometimes called implicit neural representations, are recent powerful tools for representing signals such as images with fully-connected multi-layer perceptrons (MLPs) by mapping image coordinates to their corresponding pixel values. In contrast to standard pixel-based representations, this representation is continuous with respect to network weights, allowing it to model fine detail which is limited by network capacity instead of grid resolution. Coordinate networks are commonly employed for unsupervised signal representation by training weights such that the network output has high fidelity with a single signal of interest~\cite{sitzmann2020implicit, tancik2020fourier, lindell2021bacon}. These techniques have also been demonstrated for representing medical images~\cite{wu2021irem, reed2021dynamic}. However, unsupervised approaches cannot incorporate novel high-frequency information for the task of super-resolution. 
In contrast, supervised methods learn to represent many signals over a shared function space, often in concordance with meta-learning~\cite{tancik2021learned}, learned initializations~\cite{shen2022nerp}, or convolutional structure to encode features~\cite{chen2021learning, mehta2021modulated}. These supervised methods are developed for smooth images and do not account for noise present in real-world medical imaging applications. While it's been shown that convolutional networks are prone to overfitting and hence benefit from early stopping~\cite{van2018compressed, heckel2019denoising}, the behavior of coordinate networks in noisy settings is a developing area of study. Concurrently with our work,~\citet{kim2022zero} emphasize the importance of early stopping for coordinate networks in an unsupervised setting. Tangentially, we note another concurrent work which demonstrates a scale-agnostic method for super-resolution of audio signals~\cite{kim2022learning}.
\section{Methods}

\begin{figure}[t!]
\floatconts
  {fig:diagram}
  {\caption{ \textbf{\underline{Left}: System overview}, described in Section~\ref{methods:system}. \textbf{\underline{\smash{Right}}: Image comparison} given input cropped from the red box at left. \Coord~and conv are each trained at 2$\times$. \Coord~can also be queried at 4$\times$ without re-training because it is scale-agnostic (F). \Coord~benefits from the denoising regularization term i.e.~when $\lambda \neq 0$ (D vs.~E), and \conv~does not (see Table~\ref{tab:example}).
  }}
  {\includegraphics[width=0.99\textwidth]{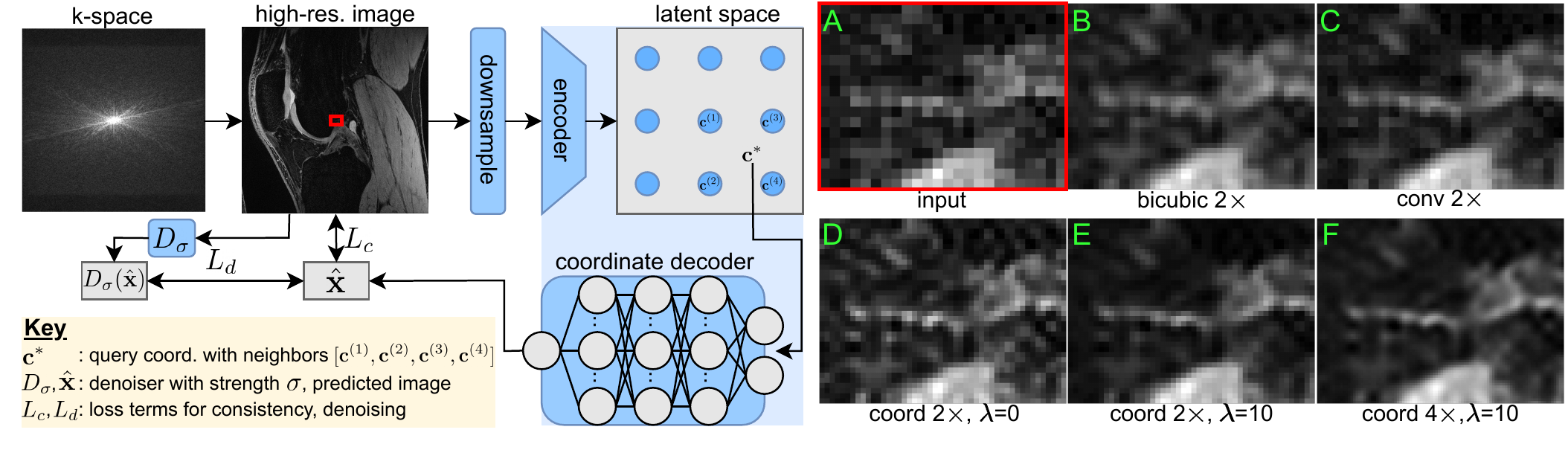}}
\end{figure}

\subsection{System Overview} \label{methods:system}
We now provide a description of our method following the diagram in Figure~\ref{fig:diagram}.
High-resolution image $\bb{x}_{hr} \in \mathbb{R}^{n \times n}$ is obtained from fully-sampled k-space measurements $\bb{y}$ characterized by the forward model $\bb{y}_i = \bb{F} \bb{S}_i \bb{x}_{hr}$, where $\bb{F}$ denotes the Fourier transform and $\bb{S}_i$ the $i^{th}$ coil sensitivity map.
Given $\bb{x}_{hr}$, a low-resolution network input $\bb{x}_{lr} \in \mathbb{R}^{\frac{n}{s} \times \frac{n}{s}}$ is created via bicubic downsampling by scale $s$. 
Subsequently an encoder $f_{\psi}: \bb{x}_{lr} \rightarrow \bb{C} \in \mathbb{R}^{d \times l \times w}$ parameterized by weights $\psi$ maps this to an equidistant feature grid.
Hence there are $l \times w$ grid points, each containing a feature vector $\bb{c} \in \mathbb{R}^d$.

These feature vectors are then input to the coordinate decoder $g_{\theta}: \mathbb{R}^d \rightarrow \mathbb{R}$ defined by:
$$ g_{\theta} \coloneqq \bb{W}_m (\alpha_{m-1} \circ \alpha_{m-2} \circ \cdots \circ \alpha_0) (\bb{c}) + \bb{b}_m, \; \; \; \; \bb{c}_j \rightarrow \alpha_j(\bb{c}_j) \coloneqq \phi(\bb{W}_j\bb{c}_j + \bb{b}_j),$$
where $\alpha_j$ corresponds to the $j^{th}$ network layer composed of weight matrix $\bb{W}_j$ and biases $\bb{b}_j$ operating on the input $\bb{c}_j$, followed by application of a nonlinear function $\phi$.

Given a particular feature vector $\bb{c}^{*} \in \mathbb{R}^d$ at a coordinate of interest, the decoder $g_{\theta}: \mathbb{R}^d \rightarrow \mathbb{R}$ queries the four neighboring latent codes $[\bb{c}^{(1)} ,\bb{c}^{(2)}, \bb{c}^{(3)}, \bb{c}^{(4)}] $
such that the decoder's output predicts the grayscale pixel value at each of those four locations. 
The pixel value at $\bb{c}^{*}$ is estimated as a linear combination of surrounding pixel values 
based on relative distance $\bb{w}^{(k)}$, i.e. $g_{\theta}(\bb{c}^{*}) = \sum_{k=1}^{4} \bb{w}^{(k)} \cdot g_{\theta}(\bb{c}^{(k)})$, consequently preventing discontinuities in the output image. 
Querying over many coordinates produces the predicted image $\hat{\bb{x}}$, so in summary we have $\hat{\bb{x}} = (g_{\theta} \circ f_{\psi})(\bb{x}_{lr})$.

We train this system end-to-end with $p$ pairs $\{\bb{x}_{lr}, \bb{x}_{hr}\}_{i=1}^p$ to find encoder and decoder weights $\psi$, $\theta$ which minimize the loss $L_c + \lambda L_d = \| \hat{\bb{x}} - \bb{x}_{hr}\|_1 + \lambda \| \hat{\bb{x}} - D_{\sigma}(\bb{x}_{hr})\|_2^2$. Here $L_c$ is a consistency loss and $L_d$ a denoising loss,
such that $D_{\sigma}$ is a denoiser with strength $\sigma$ and $\lambda$ is the relative weight of the regularization term. We use the Adam optimizer over a batch of randomly drawn image tiles for each training step and perform early stopping after $T$ iterations, as discussed below.

\subsection{Implementation}

This framework allows for many choices of encoders $f_{\psi}$ or denoisers $D_{\sigma}$. For simplicity we choose an EDSR convolutional encoder~\cite{lim2017enhanced} and BM3D denoiser~\cite{dabov2007image}, respectively. Both the denoiser strength $\sigma$ and number of training steps $T$ were tuned according to radiologist preference, resulting in $\sigma=0.03$, $T_{\coord} = 10^3$, and $T_{\conv} = 10^5$. Alternatively one could determine $T$ by maintaining a hold-out validation set, e.g.~similar to~\citet{yaman2020self}. However, this relies on quantitative metrics, which can be ambiguous, as discussed in Section~\ref{sec:results}.

We compare our decoder's continuous representation (``\textit{coord}''), to the same framework with a convolutional decoder, i.e. the original EDSR (``\textit{conv}''), which is not scale-agnostic. \Coord~is similar to \conv~but modifies the decoder to be a five-layer MLP containing 256 hidden units and ReLU activations. Each model contains roughly 1.6 million parameters and runs on a single Quadro RTX 8000 GPU. We also compare against bicubic interpolation, which can be queried at arbitrary scales but has no prior to incorporate higher frequency information.
To create the dataset, we extract 2D sagittal slices 
from SKM-TEA~\cite{desai2021skmtea}, randomly sampling five of the central 80 slices from each 3D echo-1 scan and partitioning into a 80\%/10\%/10\% train/test/validation split by volume.

\begin{figure}[b!]
\floatconts
  {fig:voxel}
  {\caption{\textbf{User interface} for Voxel, our web-based tool used in the reader study.}}
  {\includegraphics[width=0.825\textwidth]{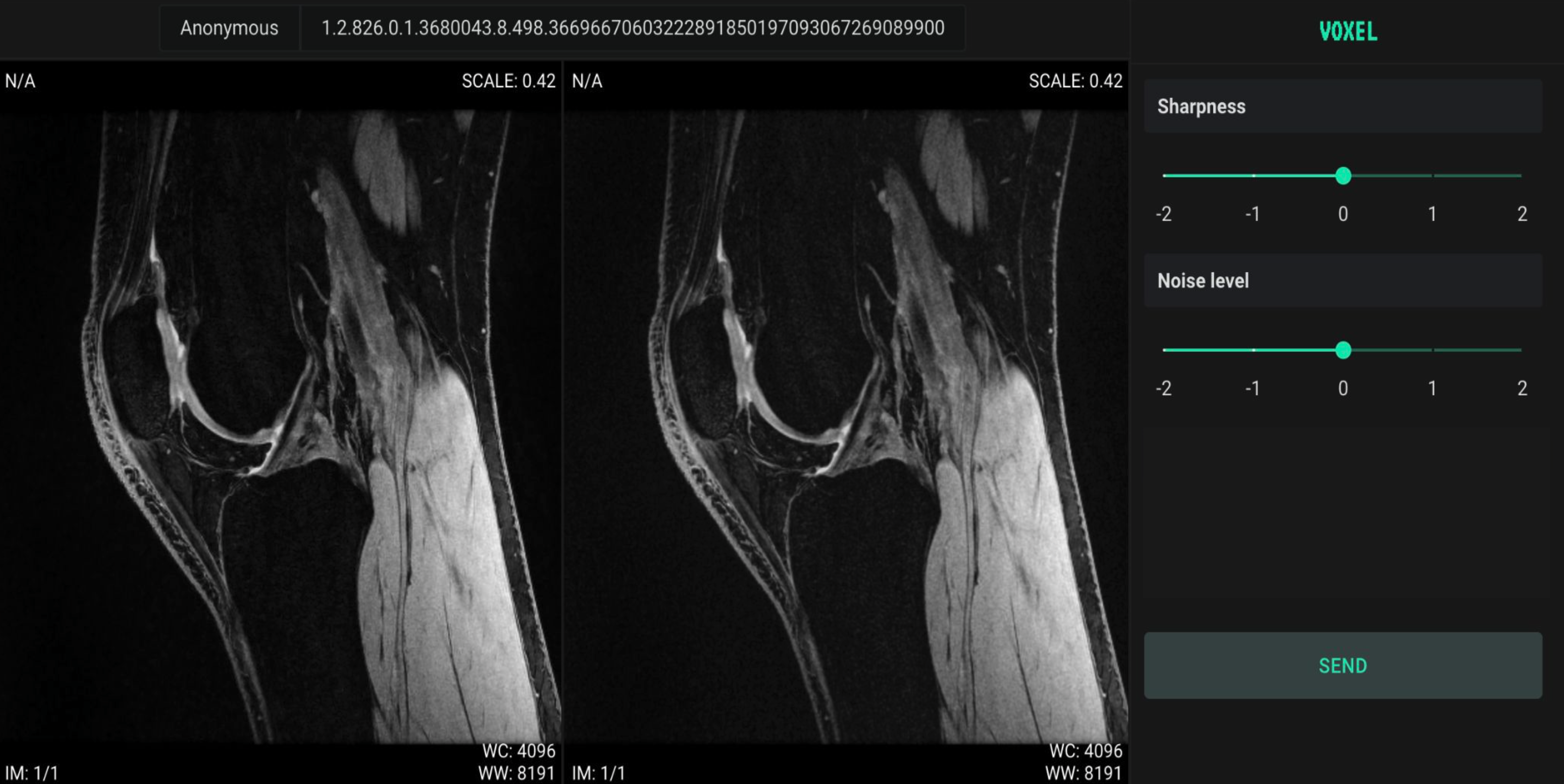}}
\end{figure}

\subsection{Reader study details} \label{methods:reader_study}
We perform a reader study with radiologists comparing \coord~and \conv, both trained on 2$\times$ super-resolution. In clinical applications one would want to scale beyond ground-truth; hence at inference we bypass downsampling and perform 2$\times$ super-resolution on the ground-truth itself. Readers used a five-point Likert scale to evaluate randomized side-by-side image pairs in terms of sharpness and noise.

The reader study was built using Voxel, a medical imaging tensor viewer for radiologist evaluation (see Fig.~\ref{fig:voxel}). This allows the reader to review images from her/his browser, forgoing the cumbersome file sharing and annotation commonly required with reader studies. Voxel provides similar features to standard DICOM viewers such as zooming, panning through image volumes, and adjusting window levels. Furthermore, responses are time-stamped, allowing for review-time analysis; while we do not investigate it here, this could enable answering questions such as, ``Would radiologists be faster diagnosing images from method A or method B?'' 
Broadly we hope this tool will enhance the impact of deep learning in medical imaging by streamlining the feedback process between researchers and radiologists. To try a demo of Voxel, please visit \url{https://voxel.im} or contact the authors.

\section{Results and Discussion}
\label{sec:results}

\subsection{Quantitative Results}
\label{sec:disc-quant}
Per Table~\ref{tab:example}, \coord~performs comparably when trained on a range of scales (1-2$\times$, row 2) vs.~a fixed scale (2$\times$, row 5). Because \coord~is scale-agnostic, it can be queried at a resolution which is both arbitrary and independent of its training scales. Conversely, \conv---without additional interpolation---queries only at fixed integer upsampling according to its training scales.
\Conv~was slightly superior to \coord~in PSNR but slightly inferior in VIF. While PSNR is biased toward smooth images, VIF is perhaps better suited to evaluate high-frequency detail augmented during super-resolution; further, it is more indicative of clinical diagnostic quality~\cite{mason2019comparison}. 

\setlength{\extrarowheight}{1.25pt}

\begin{table}[t]
\floatconts
  {tab:example}%
  {\caption{\textbf{\underline{Left}: Quantitative scores} (VIF/PSNR). \Coord~obtains similar performance to \conv~(slightly better VIF, slightly worse PSNR) and has the benefit of agnosticism with respect to training and querying scales. Below the midline we provide ablations at various training scales and values for $\lambda$, i.e.~the denoising regularization term (see discussion, Section~\ref{sec:results}).
  \textbf{\underline{\smash{Right}}: Reader study} scoring criteria (top) and results (bottom) demonstrating a slight preference for \coord.} }%
  {\includegraphics[width=0.99\textwidth]{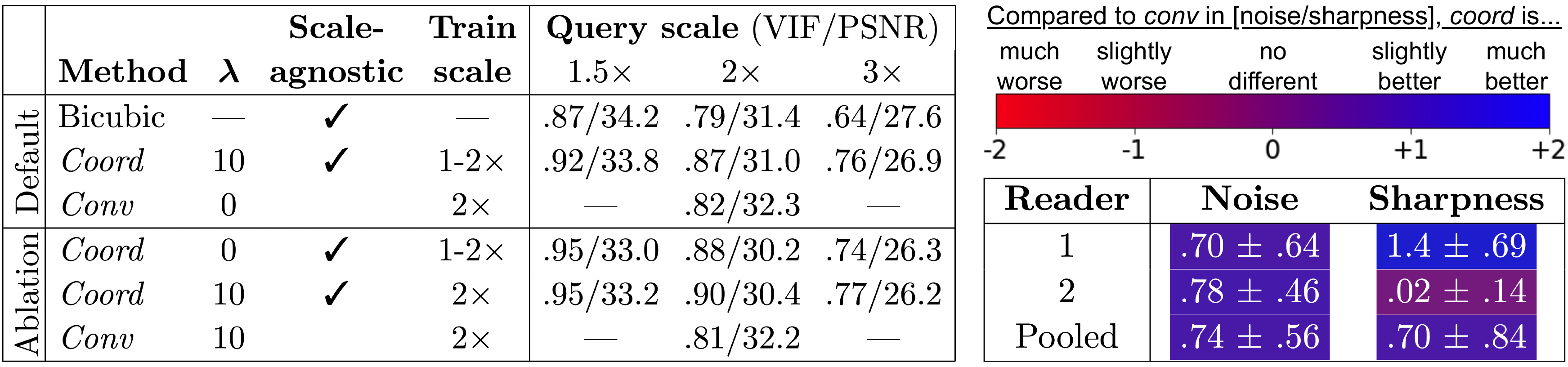}}
\end{table}
\begin{figure}[t!]
\centering
\floatconts
  {fig:early-stop-coord}
  {\caption{\textbf{Early stopping} is an important regularization technique for coordinate networks. Quantitative metrics for \textit{coord} are evaluated at each training step. Similar to the known phenomena in convolutional networks, coordinate networks overfit to high frequency components given too many training steps (see image insets), corresponding to a significant decrease in PSNR. When analyzing \textit{conv}, early stopping shows a similar benefit, albeit less severe.}}
  {\includegraphics[width=0.99\textwidth]{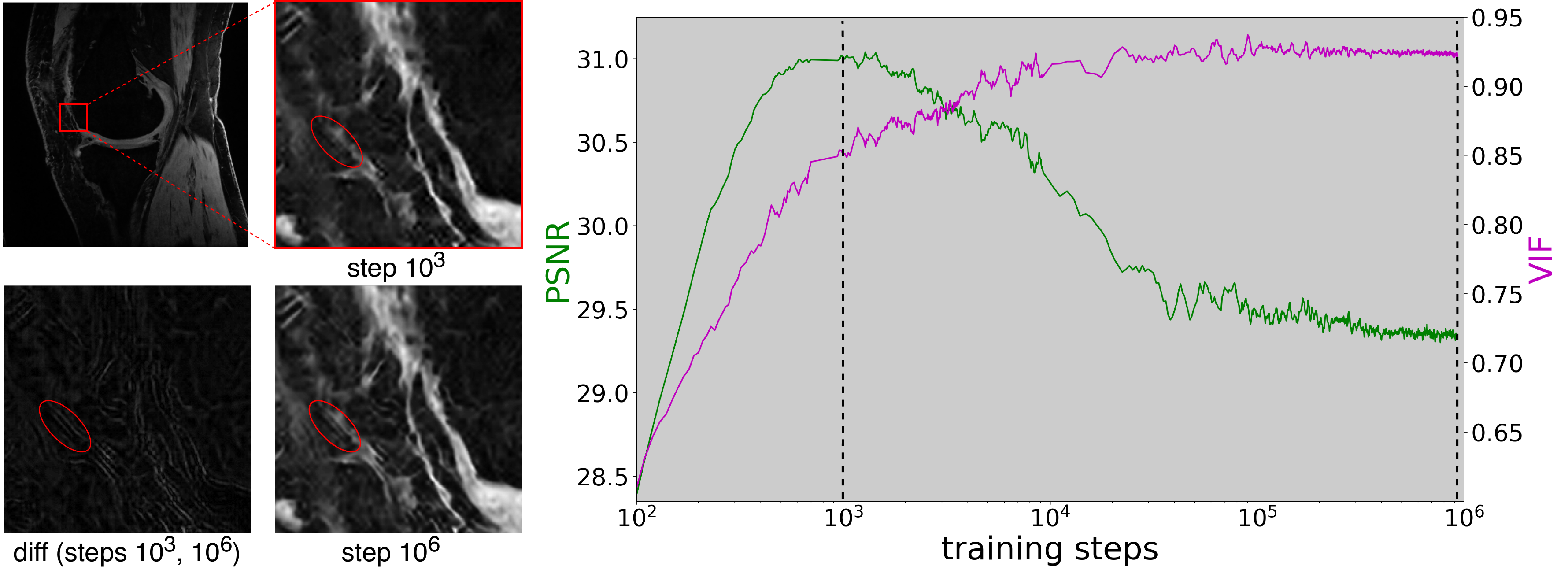}}
\end{figure}

\subsection{Reader Study}
\label{sec:disc-reader}
We note the challenge of evaluating clinical potential using image metrics alone, as they frequently do not correspond to the images preferred by end-user radiologists~\cite{kastryulin2022image}. Consider instances where higher metrics scores seemingly do not pertain to better quality: compared to \coord, $\lambda=10$ (Fig.~\ref{fig:diagram}, E), bicubic interpolation (B) achieves higher PSNR while \coord, $\lambda=0$ (D) achieves higher VIF; however, these are perceptually undesirable in terms of sharpness and noise, respectively. Furthermore, quantitative metrics require a ground-truth reference; yet in a clinical setting, the goal is to scale larger than ground-truth resolution. Hence to gain insight beyond these limitations, we present a reader study in Table~\ref{tab:example}. This demonstrates that \coord~is equivalent or slightly preferable to \conv~in terms of perceived sharpness and noise.
It's interesting to consider that despite \coord~receiving slighly lower scores on PSNR, radiologists actually prefer these images with respect to noise. Fundamentally we advocate for the use of reader studies when comparing reconstruction methods to facilitate more clinically relevant research. One tool to streamline this process is Voxel, as discussed in Section~\ref{methods:reader_study}.

\subsection{Denoising}
\label{sec:disc-noise}
Given the noise inherent to MR images, super-resolution methods such as \coord~and \conv~are prone to augment noise in the output~\cite{singh2014super}. To mitigate this effect, we leverage two techniques: denoising regularization and early stopping during network training. Unlike \coord~which benefited from denoising regularization (Fig.~\ref{fig:diagram}, D vs.~E), \conv~did not (Table~\ref{tab:example}, row 6). Meanwhile both benefit from early stopping, although the effect on \coord~(Fig.~\ref{fig:early-stop-coord}) is larger than that on \conv~(Appendix). We suspect this difference is due to the fact that convolutional structure encodes global image features, inherently acting as a denoiser. Coordinate networks do not exhibit this structural property; furthermore, their receptive field–––size of the input region that produces the output---is typically much smaller compared to convolutional networks. Overall, the comparison of coordinate and convolutional networks as image priors has not been well studied, although concurrent work has examined their behavior in unsupervised regimes~\cite{kim2022zero}. Similar to our findings, this demonstrates that coordinate networks have an inductive bias toward lower frequencies and consequently benefit from early stopping.

\section{Conclusion}

We propose a \textit{scale-agnostic} framework for MRI super-resolution to circumvent the requirement of homogeneous training data in convolutional networks. We also demonstrate various regularization strategies for coordinate networks in the presence of noise. Lastly we evaluate results using both quantitative metrics and a reader study built on Voxel.

In the future, we plan to extensively evaluate across different encoding, decoding, and denoising methods and also assess downstream impact via pixel-level quantitative MRI metrics and reader studies on various pathological regions. Crucial to this work is the behavior of coordinate networks in the presence of noise, which warrants further study both for supervised and unsupervised regimes.

\newpage
\bibliography{refs}

\newpage
\section*{Appendix}
\nopagebreak

\setcounter{figure}{0}
\makeatletter 
\renewcommand{\thefigure}{A\@arabic\c@figure}
\makeatother


\begin{figure}[b!]
\centering
    \includegraphics[width=0.99\textwidth]{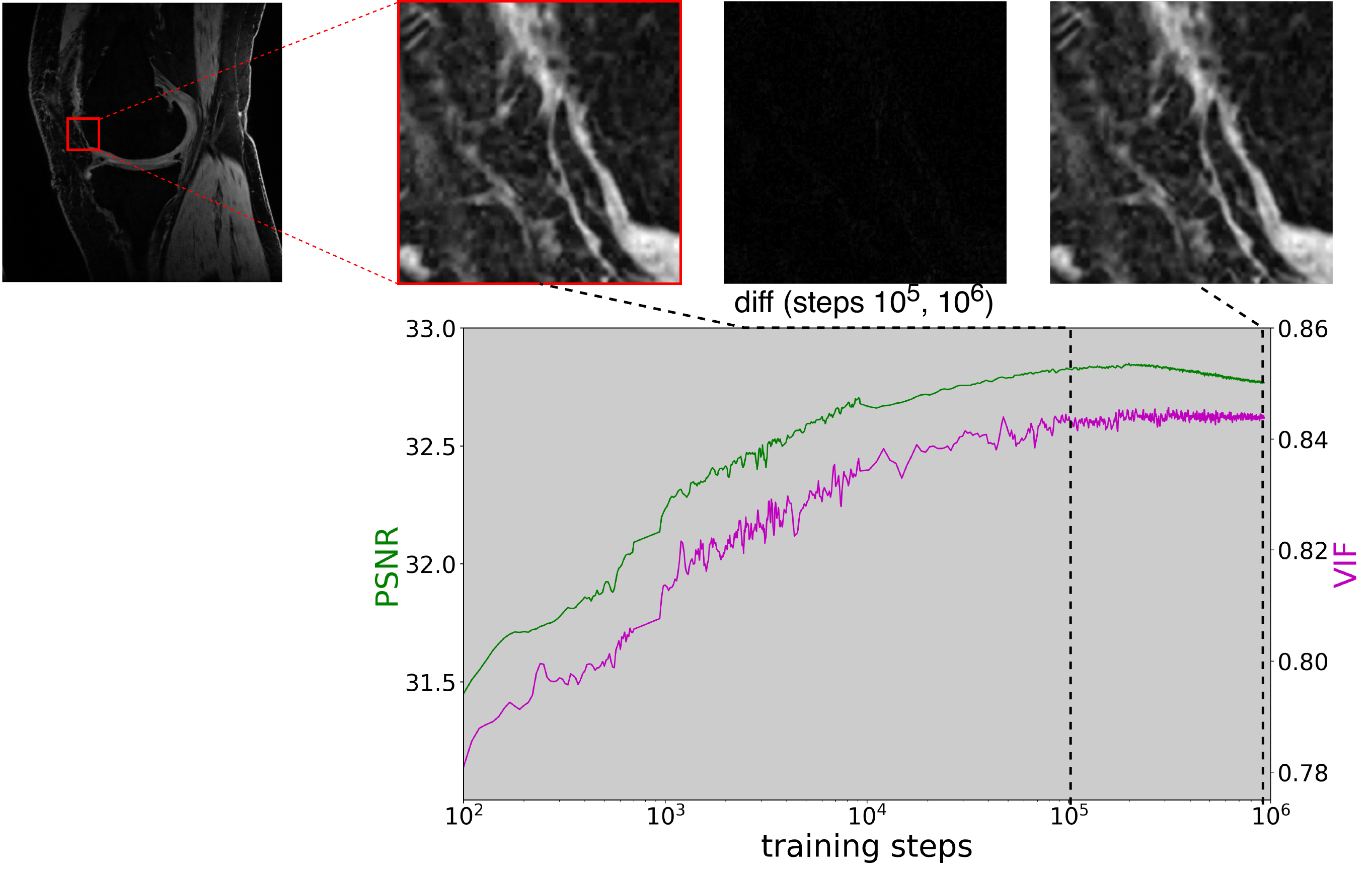}
    \caption{\textbf{Early stopping for~\conv}: Quantitative metrics are evaluated at each training step. The maximum PSNR occurs at roughly $10^5$ steps before decreasing slightly, consistent with the tendency of convolutional networks to overfit to noise. While this decrease in PSNR is non-trivial, it is less severe than the behavior of \coord, discussed in Section~\ref{sec:disc-noise}.}
    \label{fig:early-stopping-conv}
\end{figure}


\end{document}